\documentclass[lettersize, journal]{IEEEtran}

\usepackage{cite}
\usepackage{amsmath,amssymb,amsfonts}
\usepackage{algorithmic}
\usepackage{graphics}
\usepackage{hyperref}
\usepackage{textcomp}
\usepackage{xcolor}
\usepackage{siunitx}
\usepackage{tikz}
\usepackage{float}
\usepackage{ulem}
\usepackage{subcaption}
\usepackage{caption}
\usepackage{booktabs}
\usepackage{stfloats}
\usepackage{import}
\usepackage{xifthen}
\usepackage{pdfpages}
\usepackage{transparent}

\newcommand{\eg}{e.g. }

\newcommand\ddfrac[2]{\frac{\displaystyle #1}{\displaystyle #2}}

\newcommand{\vectorstyle}[1]{\boldsymbol{\mathbf{#1}}}

\begin{document}

\title{A Supervisory Learning Control Framework for Autonomous \& Real-time Task Planning for an Underactuated Cooperative Robotic task
}

\author{Sander De Witte\textsuperscript{1,2}, Tom Lefebvre\textsuperscript{1,2}, Thijs Van Hauwermeiren\textsuperscript{1,2} and Guillaume Crevecoeur\textsuperscript{1,2}\\
		\textsuperscript{1}Department of Electromechanical, Systems and Metal Engineering, Ghent University\\Tech Lane Ghent Science Park 913, B-9052 Zwijnaarde, Belgium\\
		\textsuperscript{2}Core Lab MIRO, Flanders Make Strategic Research Centre for the Manufacturing Industry
		\vspace{-25px}}

\maketitle

\begin{abstract}
	We introduce a framework for cooperative manipulation, applied on an underactuated manipulation problem. Two stationary robotic manipulators are required to cooperate in order to reposition an object within their shared work space. Control of multi-agent systems for manipulation tasks cannot rely on individual control strategies with little to no communication between the agents that serve the common objective through swarming. Instead a coordination strategy is required that queries subtasks to the individual agents. We formulate the problem in a Task And Motion Planning (TAMP) setting, while considering a decomposition strategy that allows us to treat the task and motion planning problems separately. We solve the supervisory planning problem offline using deep Reinforcement Learning techniques resulting into a supervisory policy capable of coordinating the two manipulators into a successful execution of the pick-and-place task. Additionally, a benefit of solving the task planning problem offline is the possibility of real-time (re)planning, demonstrating robustness in the event of subtask execution failure or on-the-fly task changes. The framework achieved zero-shot deployment on the real setup with a success rate that is higher than 90\%.
\end{abstract}

\section{Introduction}
\begin{figure*}[t]
	\centering
	\begin{subfigure}{0.3\textwidth}
    \fontsize{12pt}{14pt}\selectfont
    \def\svgwidth{\columnwidth}
    \scalebox{1}{
\begingroup%
  \makeatletter%
  \providecommand\color[2][]{%
    \errmessage{(Inkscape) Color is used for the text in Inkscape, but the package 'color.sty' is not loaded}%
    \renewcommand\color[2][]{}%
  }%
  \providecommand\transparent[1]{%
    \errmessage{(Inkscape) Transparency is used (non-zero) for the text in Inkscape, but the package 'transparent.sty' is not loaded}%
    \renewcommand\transparent[1]{}%
  }%
  \providecommand\rotatebox[2]{#2}%
  \newcommand*\fsize{\dimexpr\f@size pt\relax}%
  \newcommand*\lineheight[1]{\fontsize{\fsize}{#1\fsize}\selectfont}%
  \ifx\svgwidth\undefined%
    \setlength{\unitlength}{705.98027337bp}%
    \ifx\svgscale\undefined%
      \relax%
    \else%
      \setlength{\unitlength}{\unitlength * \real{\svgscale}}%
    \fi%
  \else%
    \setlength{\unitlength}{\svgwidth}%
  \fi%
  \global\let\svgwidth\undefined%
  \global\let\svgscale\undefined%
  \makeatother%
  \begin{picture}(1,0.52373988)%
    \lineheight{1}%
    \setlength\tabcolsep{0pt}%
    \put(0,0){\includegraphics[width=\unitlength,page=1]{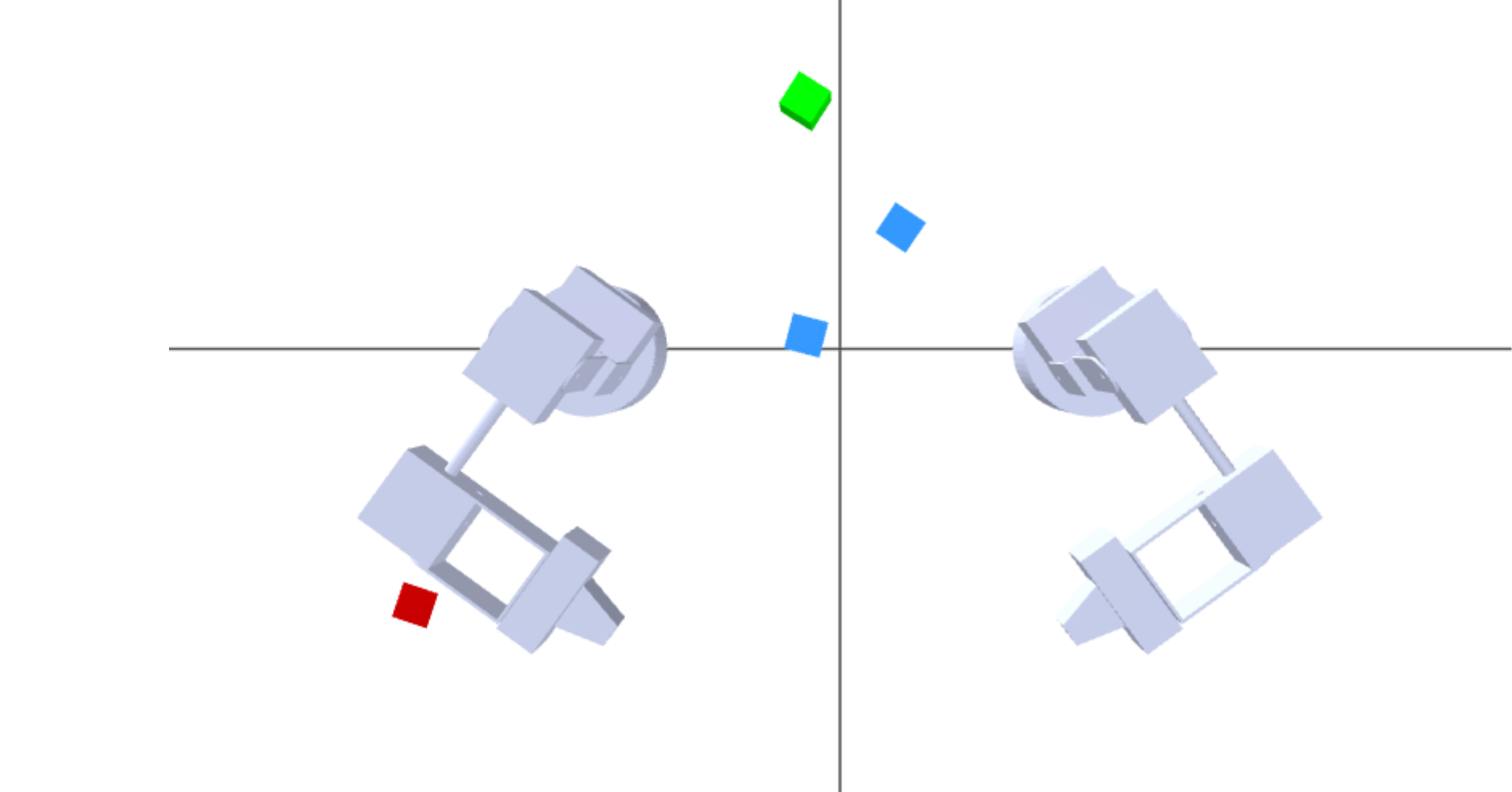}}%
    \put(0.58409334,0.22781221){\color[rgb]{0,0,0}\makebox(0,0)[t]{\lineheight{1.25}\smash{\begin{tabular}[t]{c}$(x_a,y_a)$\end{tabular}}}}%
    \put(0.3745837,0.28639261){\color[rgb]{0,0,0}\makebox(0,0)[t]{\lineheight{1.25}\smash{\begin{tabular}[t]{c}$\textrm{L}$\end{tabular}}}}%
    \put(0.73433467,0.28639261){\color[rgb]{0,0,0}\makebox(0,0)[t]{\lineheight{1.25}\smash{\begin{tabular}[t]{c}$\textrm{R}$\end{tabular}}}}%
    \put(0.64159555,0.49874826){\color[rgb]{0,0,0}\makebox(0,0)[t]{\lineheight{1.25}\smash{\begin{tabular}[t]{c}$(x_0,y_0,\theta_0)$\end{tabular}}}}%
    \put(0.27508216,0.50847959){\color[rgb]{0,0,0}\makebox(0,0)[t]{\lineheight{1.25}\smash{\begin{tabular}[t]{c}$\begin{cases}f_1=0\\b_1 = \textrm{L}\end{cases}$\end{tabular}}}}%
    \put(0,0){\includegraphics[width=\unitlength,page=2]{schematic_firstaction.pdf}}%
  \end{picture}%
\endgroup%
}

	  	\caption{First action $a_1=(0,L,x_a,y_a)$}
	  	\label{subfig:action1}
	\end{subfigure}
	\begin{subfigure}{0.3\textwidth}
    \fontsize{12pt}{14pt}\selectfont
    \def\svgwidth{\columnwidth}
    \scalebox{1}{
\begingroup%
  \makeatletter%
  \providecommand\color[2][]{%
    \errmessage{(Inkscape) Color is used for the text in Inkscape, but the package 'color.sty' is not loaded}%
    \renewcommand\color[2][]{}%
  }%
  \providecommand\transparent[1]{%
    \errmessage{(Inkscape) Transparency is used (non-zero) for the text in Inkscape, but the package 'transparent.sty' is not loaded}%
    \renewcommand\transparent[1]{}%
  }%
  \providecommand\rotatebox[2]{#2}%
  \newcommand*\fsize{\dimexpr\f@size pt\relax}%
  \newcommand*\lineheight[1]{\fontsize{\fsize}{#1\fsize}\selectfont}%
  \ifx\svgwidth\undefined%
    \setlength{\unitlength}{707.66221367bp}%
    \ifx\svgscale\undefined%
      \relax%
    \else%
      \setlength{\unitlength}{\unitlength * \real{\svgscale}}%
    \fi%
  \else%
    \setlength{\unitlength}{\svgwidth}%
  \fi%
  \global\let\svgwidth\undefined%
  \global\let\svgscale\undefined%
  \makeatother%
  \begin{picture}(1,0.52249507)%
    \lineheight{1}%
    \setlength\tabcolsep{0pt}%
    \put(0,0){\includegraphics[width=\unitlength,page=1]{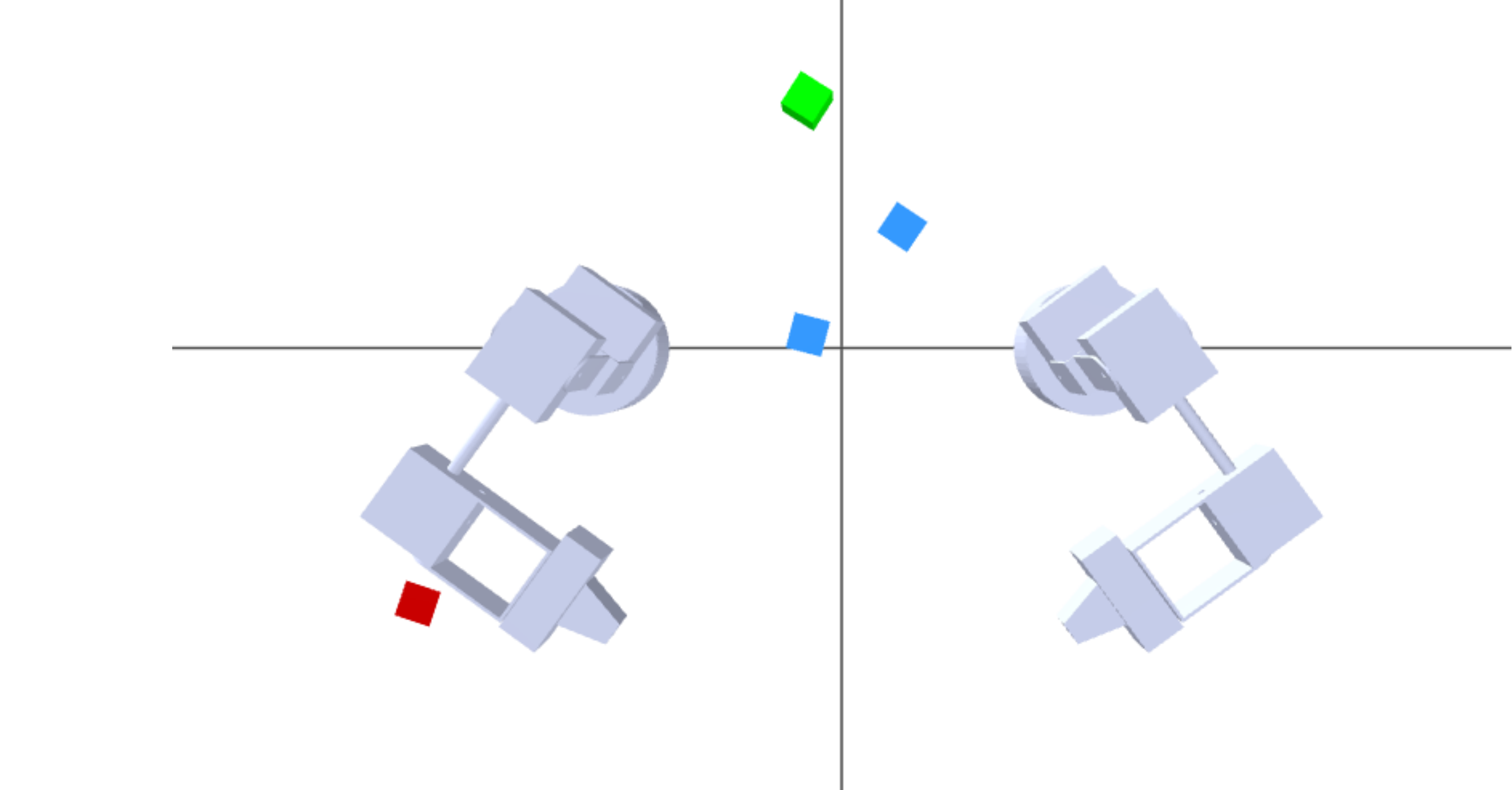}}%
    \put(0.6364814,0.41917832){\color[rgb]{0,0,0}\makebox(0,0)[t]{\lineheight{1.25}\smash{\begin{tabular}[t]{c}$(x_b,y_b)$\end{tabular}}}}%
    \put(0.37449632,0.28665108){\color[rgb]{0,0,0}\makebox(0,0)[t]{\lineheight{1.25}\smash{\begin{tabular}[t]{c}$\textrm{L}$\end{tabular}}}}%
    \put(0.73481934,0.28665108){\color[rgb]{0,0,0}\makebox(0,0)[t]{\lineheight{1.25}\smash{\begin{tabular}[t]{c}$\textrm{R}$\end{tabular}}}}%
    \put(0.42905606,0.20240373){\color[rgb]{0,0,0}\makebox(0,0)[t]{\lineheight{1.25}\smash{\begin{tabular}[t]{c}$(x_1,y_1,\theta_1)$\end{tabular}}}}%
    \put(0.27580007,0.50727105){\color[rgb]{0,0,0}\makebox(0,0)[t]{\lineheight{1.25}\smash{\begin{tabular}[t]{c}$\begin{cases}f_1=0\\b_1 = \textrm{R}\end{cases}$\end{tabular}}}}%
    \put(0,0){\includegraphics[width=\unitlength,page=2]{schematic_secondaction.pdf}}%
  \end{picture}%
\endgroup%
}

	  	\caption{Second action $a_2=(0,R,x_b,y_b)$}
	  	\label{subfig:action2}
	\end{subfigure}
	\begin{subfigure}{0.3\textwidth}
    \fontsize{12pt}{14pt}\selectfont
    \def\svgwidth{\columnwidth}
    \scalebox{1}{
\begingroup%
  \makeatletter%
  \providecommand\color[2][]{%
    \errmessage{(Inkscape) Color is used for the text in Inkscape, but the package 'color.sty' is not loaded}%
    \renewcommand\color[2][]{}%
  }%
  \providecommand\transparent[1]{%
    \errmessage{(Inkscape) Transparency is used (non-zero) for the text in Inkscape, but the package 'transparent.sty' is not loaded}%
    \renewcommand\transparent[1]{}%
  }%
  \providecommand\rotatebox[2]{#2}%
  \newcommand*\fsize{\dimexpr\f@size pt\relax}%
  \newcommand*\lineheight[1]{\fontsize{\fsize}{#1\fsize}\selectfont}%
  \ifx\svgwidth\undefined%
    \setlength{\unitlength}{707.40275416bp}%
    \ifx\svgscale\undefined%
      \relax%
    \else%
      \setlength{\unitlength}{\unitlength * \real{\svgscale}}%
    \fi%
  \else%
    \setlength{\unitlength}{\svgwidth}%
  \fi%
  \global\let\svgwidth\undefined%
  \global\let\svgscale\undefined%
  \makeatother%
  \begin{picture}(1,0.52268671)%
    \lineheight{1}%
    \setlength\tabcolsep{0pt}%
    \put(0,0){\includegraphics[width=\unitlength,page=1]{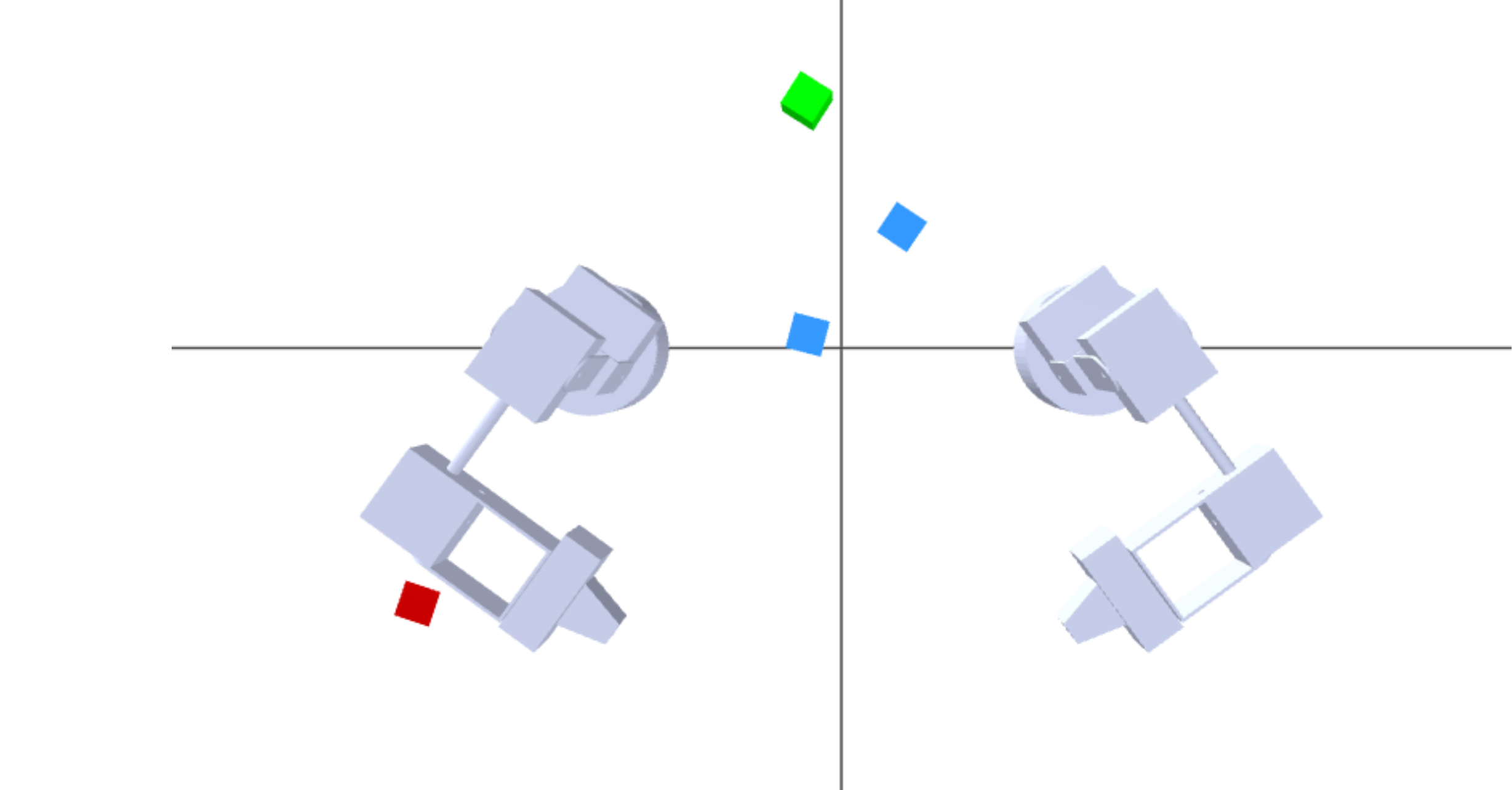}}%
    \put(0.61461648,0.42938627){\color[rgb]{0,0,0}\makebox(0,0)[t]{\lineheight{1.25}\smash{\begin{tabular}[t]{c}$(x_2,y_2,\theta_2)$\end{tabular}}}}%
    \put(0.37415922,0.28557627){\color[rgb]{0,0,0}\makebox(0,0)[t]{\lineheight{1.25}\smash{\begin{tabular}[t]{c}$\textrm{L}$\end{tabular}}}}%
    \put(0.73452085,0.28557627){\color[rgb]{0,0,0}\makebox(0,0)[t]{\lineheight{1.25}\smash{\begin{tabular}[t]{c}$\textrm{R}$\end{tabular}}}}%
    \put(0.22040137,0.02836388){\color[rgb]{0,0,0}\makebox(0,0)[t]{\lineheight{1.25}\smash{\begin{tabular}[t]{c}$(x_g,y_g,\theta_g)$\end{tabular}}}}%
    \put(0.27452901,0.50745711){\color[rgb]{0,0,0}\makebox(0,0)[t]{\lineheight{1.25}\smash{\begin{tabular}[t]{c}$\begin{cases}f_1=1\\b_1 = \textrm{L}\end{cases}$\end{tabular}}}}%
    \put(0,0){\includegraphics[width=\unitlength,page=2]{schematic_thirdaction.pdf}}%
  \end{picture}%
\endgroup%
}

	  	\caption{Final action $a_3=(1,L,x_g,y_g)$}
	  	\label{subfig:action3}
	\end{subfigure}
	\caption{Schematic figure showing multiple collaborative actions to pick-and-place object to its desired orientation, which is unachievable by an individual manipulator because it has only five DoFs.}
	\label{fig:schematic}
\end{figure*}

Cooperative robots are an increasingly important aspect of modern robotics, with applications ranging from unmanned vehicles to robotic manipulators \cite{Kim2018, Ta2020, Viegas2017}. Cooperative robots are able to work together to serve a common goal, leveraging their combined capabilities to perform tasks that would be impossible for a single robot to accomplish. For example, individual agents are limited in performing an assembly task on their own; these limitations can be overcome by using a cooperative set-up of robotic manipulators \cite{Dogar2019}. Cooperation results in the ability to perform tasks with a higher complexity, by increasing the manipulation space of a single robot by using multiple robots. In this study, we examine how two robotic manipulators can collaborate to complete complex tasks by sharing a portion of their individual workspace. The manipulators are tasked with moving an object to a desired location and orientation within the global workspace. By working together, the manipulators can achieve these goals through multiple, collaborative actions in the shared workspace, illustrated in Fig. \ref{fig:schematic}.

Task and Motion Planning encompass the solving of combined problems involving a finite sequence of discrete mode types, \eg which objects to pick and place, continuous mode parameters, the poses and grasps associated with the movable objects, and the continuous motion paths \cite{Garrett2020}. Standard TAMP approaches perform a combined optimization of all these parameters, handling the interdependence of the motion-level and the task-level aspects of the problem. Cooperation between different robots performing a single task, can be described as a TAMP problem. Which robot to query to perform a certain action is then an example of an extra discrete mode type. In addition we can look at the above mentioned cooperation problem as such a problem.
In this approach however, the problem is decomposed in separate scheduling and motion planning problems. The discrete decisions are completely disconnected from the continuous motion. The next best task, a new pick-and-place action with its associated grasp poses for one of the robots, is determined without performing a trajectory optimization for each action, resulting in a simplified optimization problem. 

To cope with changes in the environment, such as changes in the desired goal configuration or difficulties with individual subtasks, we aim to develop a parametrized policy, to determine the next best action, that only depends on the current object configuration and the target goal configuration. The policy will determine the optimal sequence of actions, including which manipulator should perform each manipulation, to reach the desired goal. Real-time computation benefits from this control decomposition in supervisory control and low-level control, as the associated optimization problem is significantly reduced.

The found policy could be used to accelerate interleaved TAMP approaches as well, as is shown in other research, that apply learning techniques to accelerate TAMP. Driess et al. learn a classifier to determine the feasibility of the resulting geometric problem for a discrete decision based on an image of the workspace \cite{Driess2020}. In the research by Chitnis et al. \cite{Chitnis2016} learning methods are used to guide a combined search to find high-level symbolic plans and their low-level refinements, \eg the associated motion plans and grasp poses. At the low-level they learn how to refine the high-level plans, by learning to propose continuous values for symbolic references that are likely to result in collision-free trajectories. In a similar vein, our method could be used to output the next best action that has a feasible trajectory, thus guiding the search for an optimal action sequence.

Our approach uses a reinforcement learning approach to train this parametrized policy. Reinforcement learning is a machine learning technique that allows agents to learn from their experiences and improve their decision-making over time. Deep reinforcement learning has led to a number of successes in learning policies for sequential decision-making problems in both simulated environments \cite{Schrittwieser2020,Mnih2015} as robotic tasks \cite{Nguyen2019,Singh2019}. Translation of policies trained in simulation to the real world poses a major problem of current RL research in robotics. The latter forms less of a problem in our approach, since the learning is performed on a higher level.

Using a discretized version of the shared workspace, from which the next best action is chosen works \cite{DeWitte2022}, but results in a low success rate and is difficult to train. This could be circumvented by using a continuous approximation of the shared workspace, resulting in a more complete parametrization of the shared workspace. The parametrization is no longer limited to a set of discrete points in the shared workspace, but comprises it completely. The motivating question is if the continuous parametrization results in better results, such as an increased training speed and a better success rate.

\section{Problem statement} \label{sec:problem-statement}
We start by describing the small-scale setup and its associated manipulation problem. The setup consist out of a work cell equipped with two low-cost plane mirrored manipulators and a perception system. The perception system consist of a camera attached at the top of the work cell with a clear top-down view of the workspace and is able to identify and locate objects. Both manipulators have five degrees of freedom (DoFs) and are equipped with a low-level controller, capable of pick-and-placing an object in the base $xy$-plane. This plane will be referred to as the manipulation plane.

Two DoFs are used to control the pitch and roll of the end-effector, during pick-up and drop-off these need to be zero to ensure a correct grasp and release. Two DoFs are used to control the position in the manipulation plane, while the last DoF is used to control the height. Therefore, we can not directly control the yaw of the robot's end-effector and consequently the yaw of the object to manipulate, henceforth simply referred to as the object's orientation. This results in an underactuated manipulation problem.

The final orientation will be the result of consecutive pick-and-place maneuvers performed by each robot, more specifically the pick-up and drop-off positions and the geometry of the manipulators. The individual manipulation planes of the left and right (respectively L and R) robots are referred to as $\mathcal{M}_{\textrm{L}} \subset\mathbb{R}^2$ and $\mathcal{M}_{\textrm{R}} \subset\mathbb{R}^2$, which are defined as the controllable set of end-effector configurations in the manipulation plane expressed as Cartesian coordinates with respect to a global frame of reference. In other words, the individual manipulation spaces are determined as the intersection of the individual manipulator workspaces, expressed in the global frame of reference, and the manipulation plane. These feasible workspaces consist out of the collection of points that are collision-free, including self-collision and collision with the ground. Additionally we can find the global $\mathcal{M}$ and the mutual $\mathcal{M}_m$ manipulation spaces, respectively, as the union and intersection of the individual workspaces.
\begin{align*}
	\mathcal{M} &= \mathcal{M}_\text{L} \cup \mathcal{M}_\text{R}  \\
	\mathcal{M}_m &= \mathcal{M}_\text{L} \cap \mathcal{M}_\text{R} 
\end{align*}

A pick-and-place task is considered where objects, placed in the global manipulation space, need to be repositioned to a new and given goal position $(x_g,y_g) \in \mathcal{M}$ and desired orientation $\theta_g$. By adding this orientation to the manipulation space, the state space $\mathcal{S}=\mathcal{M}\times\mathbb{R}$ is obtained. A successful manipulation is achieved if the object has the desired position and orientation. When for example the goal position $\vectorstyle{x}_g \in\mathcal{M}_\text{L}$ and the present position $\vectorstyle{x}_t \in \mathcal{M}_\text{R}$, it is clear that the task can only be achieved by a hand over in the mutual work space. Furthermore considering that the control of the individual manipulators is such that they can not control the orientation of the object, a sequence of hand-overs will be required to maneuver the desired object into its desired configuration $\vectorstyle{s}_g = (x_g,y_g,\theta_g) \in \mathcal{S} = \mathcal{M} \times \mathbb{R}$. In particular a sequence of varying hand-over positions must be determined in a way that the subsequent relative reorientations accumulate into the desired final orientation. 

The objective of this work is therefore to find a control strategy capable of querying a sequence of isolated pick-and-place maneuvers, each of which can be executed by a single manipulator, and, so that such a sequence exists from any arbitrary starting configuration to any desired goal configuration that intersects with the global manipulation space.

\section{Methodology}
The control problem described in section \ref{sec:problem-statement} qualifies as a TAMP problem since both the optimal hand-over sequence can be determined as well as the individual motions in between each hand-over. In this work we propose a hierarchical control strategy decomposing the TAMP problem into separate scheduling and motion planning problems.

\subsection{Mathematical problem definition}
The state of the system, indicated by $\vectorstyle{s}_t$, consist out of the Cartesian coordinates of the manipulated object and its orientation. Similarly, the goal configuration of the object is denoted as $\vectorstyle{s}_g$.
\begin{align*}
	\vectorstyle{s}_t &= (x_t,y_t,\theta_t) \in \mathcal{S} \\
	\vectorstyle{s}_g &= (x_g,y_g,\theta_g) \in \mathcal{S}
\end{align*}

The objective is to design a policy function $\pi:\mathcal{S}^2\mapsto\mathcal{A}$, so that for any initial state in the state space $s_0 \in \mathcal{S}$, a finite sequence of pick-and-place actions $\vectorstyle{a}_t \in \mathcal{A}$ can be determined such that the object can be manipulated into any desired goal configuration in the observation space $\vectorstyle{s}_g \in \mathcal{S}$. We will detail the definition of the action in the next paragraph. For the moment it is assumed that the action contains sufficient information for the system to change the state of the object from $\vectorstyle{s}_t$ to $\vectorstyle{s}_{t+1}$. This problem can be then formulated as a first-exit optimal control problem.
\begin{equation}
	\label{eq:problem1}
	\pi^* = \arg \max_\pi \underset{\vectorstyle{s}_0\sim\mathcal{U}(\mathcal{S})}{\mathbb{E}} \left[\sum\nolimits_{t=0}^{t'} r(\vectorstyle{s}_t,\vectorstyle{s}_g,{\pi}(\vectorstyle{s}_t,\vectorstyle{s}_g))\right]
\end{equation}
Here $\mathcal{U}(\mathcal{S})$ denotes the uniform distribution on $\mathcal{S}$.
The reward $r:\mathcal{S}^2\times\mathcal{A}\mapsto\mathbb{R}$ can be used to express different abstract features of the policy such as maximum accuracy or minimal execution time or can directly relate to the underlying motion optimization problem. We discuss this possibility in the final paragraph. Finally, since a first exit-optimal control problem is considered with a time independent reward function, the optimal policy function will be time-invariant.

Let us now characterize the information that is encoded in to the action, $\vectorstyle{a}_t$. The action consists out of three separate decisions:
\begin{enumerate}
	\item whether the present action is final, indicated by the Boolean $f_t\in{0,1}$;
	\item whether the left or right manipulator needs to be queried, given by the discrete variable $b_t\in\{\text{L},\text{R}\}$;
	\item the next drop-off position, represented by $(x_{t+1},y_{t+1}) \in \mathcal{M}$ in Cartesian coordinates.
\end{enumerate}

Formally, the following action space is retrieved:
\begin{equation}
	\label{eq:action}
	\vectorstyle{a}_t = (f_t,b_t,x_{t+1},y_{t+1})\in \mathcal{A}=\{0,1\}\times \{\text{L},\text{R}\}\times\mathcal{M}
\end{equation}
Presented with this information the selected manipulator should be able to determine and execute the motion $(x_t,y_t)\rightarrow(x_{t+1},y_{t+1})$. Some further rationalizations can be imposed on the policy. In practice the drop-off position is limited to the mutual manipulation space $(x_t,y_t)\in\mathcal{M}_m$ when $f_t=0$, as this is the only space both agents can reach. Only in the final step, $f_t=1$, the action can coincide with the global manipulation space, as the drop-off position should coincide with the goal position.
Which agent to query is not straightforward, since the goal state's location can be anywhere in the global manipulation space. Though it may seem rational to alternate between the two manipulators in between every query, taking into account the possibility that the manipulator does not successfully execute the motion $(x_t,y_t)\rightarrow(x_{t+1},y_{t+1})$. We opt to leave this decision to the policy rather than imposing alternation. Additionally the goal configuration could also change mid-execution.

The state update rule is determined by a nonlinear function, that depends on the forward kinematics of the queried robotic manipulator, and does not depend on the motion plan between two positions, since the relative orientation between the object and the robot is assumed to be constant.
\begin{align*}
	\vectorstyle{s}_{t+1} & = \vectorstyle{f}(\vectorstyle{s}_t,\vectorstyle{s}_g,\vectorstyle{a}_t) \\
	& =(x_{t+1}(\vectorstyle{s}_g,\vectorstyle{a}_t),y_{t+1}(\vectorstyle{s}_g,\vectorstyle{a}_t),\theta(\vectorstyle{s}_t,\vectorstyle{s}_g,\vectorstyle{a}_t))
\end{align*}

In conclusion we may note that the low level motion plan, i.e. continuous manipulator path between the two positions and $(x_t,y_t)\rightarrow(x_{t+1},y_{t+1})$, is determined and performed by the individual manipulators themselves. It is also possible that the motion plan is optimal and maximizes some reward. In the most general case this means that the scheduling reward can be represented as an underlying path planning problem. 
\begin{equation}
	\label{eq:problem2}
	\begin{aligned}
		r(\vectorstyle{s}_t,\vectorstyle{s}_g,\vectorstyle{a}_t) &= \min_{\vectorstyle{q}(\tau)} \int_0^1 c(\vectorstyle{q}(\tau),\vectorstyle{s}_g) \textit{d}\tau \\
		\text{s.t. } \vectorstyle{p}_t &= \mathcal{F}_{b_t}(\vectorstyle{q}(0)) \\
		\vectorstyle{p}_{t+1} &= \mathcal{F}_{b_t}(\vectorstyle{q}(1)) 
	\end{aligned}
\end{equation}

A full TAMP approach would thus require to solve problem (\ref{eq:problem1}) and (\ref{eq:problem2}) simultaneously. In this work we decouple both problems completely, neglecting the possible influence of the motion planning problem on the task scheduling. However for this influence to have any significant effect on the solution, the motion planning objective should be represented in the scheduling problem. In this work we focus on sparse rewards, hence validating the assumption.

\begin{figure}
	\centering
    \fontsize{12pt}{14pt}\selectfont
    \def\svgwidth{\textwidth}
    \scalebox{0.65}{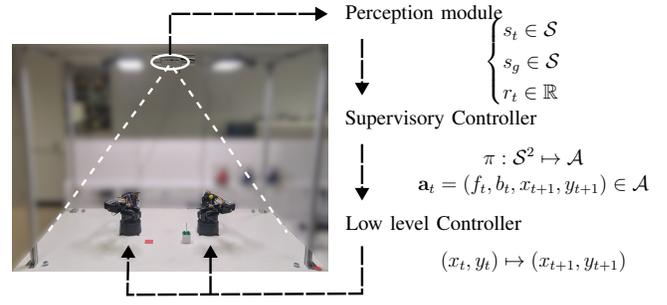}

	\caption{Overview of the method, showing clear decomposition between supervisory controller and low level controller.}
	\label{fig:overview}
\end{figure}

\subsection{Solution strategy}
In this section we discuss how we solve problem (\ref{eq:problem1}). Our solution approach requires us to parametrize the action and thus in particular the mutual manipulation space, $\mathcal{M}$. Second we discuss a parametrization strategy for the optimal policy, $\pi^*$. Finally we discuss a strategy to find the optimal policy by learning optimal parameters.

\subsubsection{\textbf{Parametrization of the solution space}}
\begin{figure}[ht]
    \centering
    \fontsize{12pt}{14pt}\selectfont
    \def\svgwidth{\columnwidth}
    \scalebox{0.9}{
\begingroup%
  \makeatletter%
  \providecommand\color[2][]{%
    \errmessage{(Inkscape) Color is used for the text in Inkscape, but the package 'color.sty' is not loaded}%
    \renewcommand\color[2][]{}%
  }%
  \providecommand\transparent[1]{%
    \errmessage{(Inkscape) Transparency is used (non-zero) for the text in Inkscape, but the package 'transparent.sty' is not loaded}%
    \renewcommand\transparent[1]{}%
  }%
  \providecommand\rotatebox[2]{#2}%
  \newcommand*\fsize{\dimexpr\f@size pt\relax}%
  \newcommand*\lineheight[1]{\fontsize{\fsize}{#1\fsize}\selectfont}%
  \ifx\svgwidth\undefined%
    \setlength{\unitlength}{1327.98695316bp}%
    \ifx\svgscale\undefined%
      \relax%
    \else%
      \setlength{\unitlength}{\unitlength * \real{\svgscale}}%
    \fi%
  \else%
    \setlength{\unitlength}{\svgwidth}%
  \fi%
  \global\let\svgwidth\undefined%
  \global\let\svgscale\undefined%
  \makeatother%
  \begin{picture}(1,0.60299464)%
    \lineheight{1}%
    \setlength\tabcolsep{0pt}%
    \put(0,0){\includegraphics[width=\unitlength,page=1]{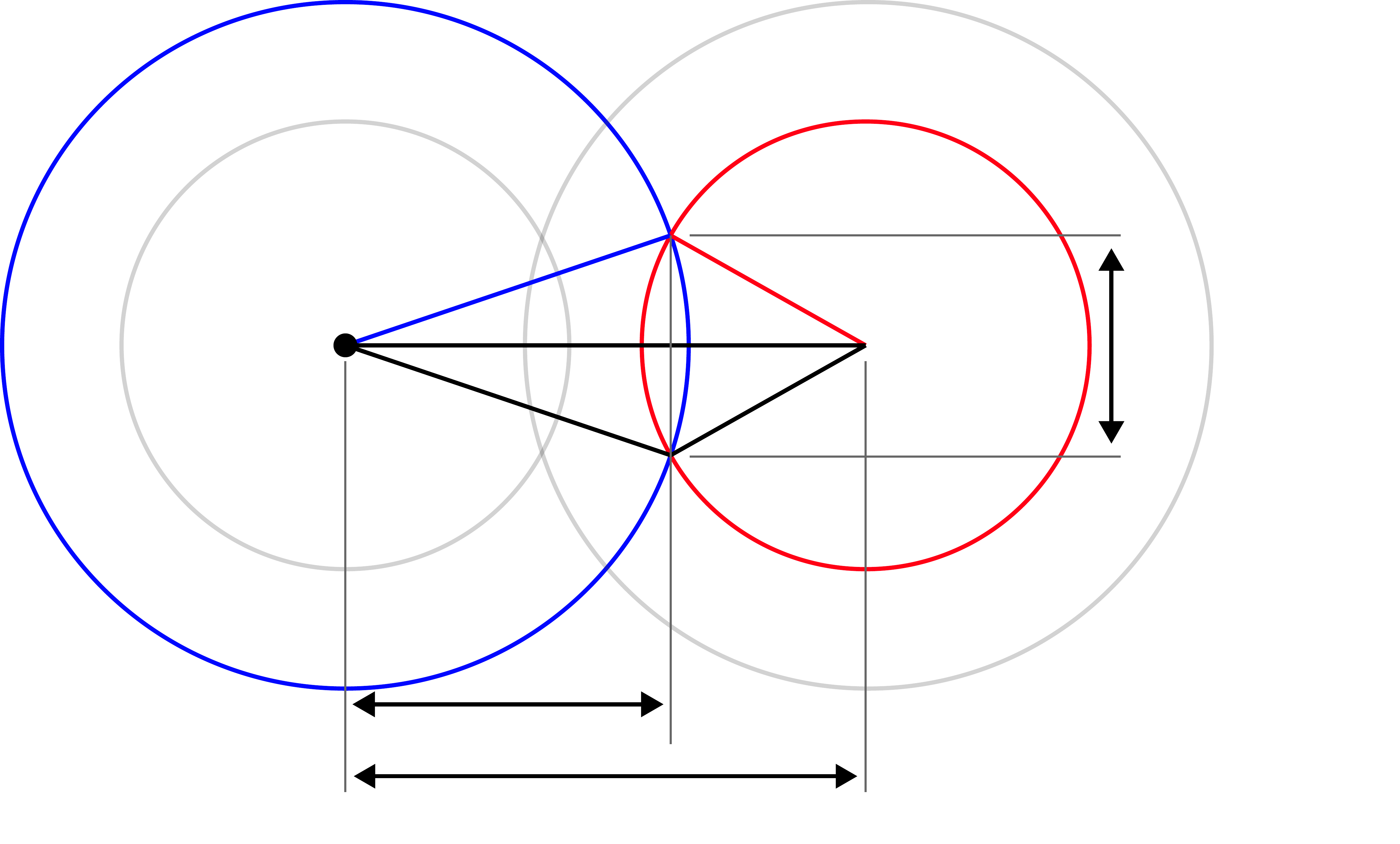}}%
    \put(0.36291722,0.06752035){\color[rgb]{0,0,0}\makebox(0,0)[t]{\lineheight{1.25}\smash{\begin{tabular}[t]{c}$y$\\\end{tabular}}}}%
    \put(0.43250198,0.00637124){\color[rgb]{0,0,0}\makebox(0,0)[t]{\lineheight{1.25}\smash{\begin{tabular}[t]{c}$d$\\\end{tabular}}}}%
    \put(0.87837289,0.34505479){\color[rgb]{0,0,0}\makebox(0,0)[t]{\lineheight{1.25}\smash{\begin{tabular}[t]{c}$a=2\cdot x$\\\end{tabular}}}}%
    \put(0,0){\includegraphics[width=\unitlength,page=2]{cont_repr.pdf}}%
    \put(0.13363234,0.32825324){\color[rgb]{0,0,0}\makebox(0,0)[t]{\lineheight{1.25}\smash{\begin{tabular}[t]{c}$y$\\\end{tabular}}}}%
    \put(0.22590238,0.37764107){\color[rgb]{0,0,0}\makebox(0,0)[t]{\lineheight{1.25}\smash{\begin{tabular}[t]{c}$(0,0)$\\\end{tabular}}}}%
    \put(0.07011941,0.38024746){\color[rgb]{0,0,0}\makebox(0,0)[t]{\lineheight{1.25}\smash{\begin{tabular}[t]{c}$x$\\\end{tabular}}}}%
  \end{picture}%
\endgroup%
}

    \caption{Visual representation of the continuous representation.}
    \label{fig:cont_repr}
\end{figure}

As mentioned previously, we can limit the drop-off position to the mutual manipulation space. The feasible positions in the manipulation plane for each manipulator lie in between two concentric circles, with a minimum radius and maximum radius, $\underline{r}$ and $\overline{r}$ respectively. Since the set-up is plane symmetric (or mirrored), these are the same for each robot. This is visualized in Fig. \ref{fig:cont_repr}.

We can parametrize the mutual workspace with two variables, $\alpha\in [-1,1]$ and $\beta\in [-1,1]$, representing the $x$- and $y$-position respectively. We can normalize the $x$-position between its maximum, $\overline{x}$, and minimum, $\underline{x}$, which is the intersection of the two large circles. Since $\underline{x} = - \overline{x}$ we can scale $\alpha$ with this maximum value to retrieve $x$.
$$x=\alpha\cdot \overline{x}$$
with
$$\overline{x} = \sqrt{\overline{r}^2-d^2/4}$$

Moving up from $\underline{x}$ to $\overline{x}$, the width of the mutual workspace changes. Therefore the maximum and minimum $y$-value, respectively $\overline{y}$ and $\underline{y}$, vary with $x$. The width either correspond to the smaller, inside the bounds $[x',-x']$, or larger circle, outside the bounds. The value where the width changes can be found as the intersection between the blue and red circle, as shown in Fig. \ref{fig:cont_repr}.

$$x' = \sqrt{\ddfrac{4d^2R^2-(d^2-r^2+R^2)^2}{4d^2}}$$
Given a certain $x$, $y$ can be retrieved from $\beta$ as:
$$y=\beta\cdot \overline{y}$$
with
\begin{equation*}
	\overline{y} = \left\lbrace
	\begin{aligned}
	\sqrt{\underline{r}^2-|x|^2},& && x \in [-x',x'] \\
	\sqrt{\overline{r}^2-|x|^2},& && x \in [-\overline{x},-x'] \cup [x',\overline{x}]
	\end{aligned}\right.
\end{equation*}
and assuming again, due to the symmetry, that $\underline{y}=-\overline{y}$.

\subsubsection{\textbf{Policy definition}} \label{sec:policy-def}
We opt to find a parametrized policy $\pi(\cdot,\cdot) \approx \pi(\cdot,\cdot;\vectorstyle{\phi})$, which assigns a subtask each discrete time step to one of the manipulators. These perform their own low-level motion plan based on this given subtask. This means that instead of solving problem (\ref{eq:problem1}) for a specific initial state each time, we aim to solve for a global policy representation. This has the advantage that once the policy has been learning, the policy can be queried explicitly saving essential computational resources in a real-time setting.

Since the policy is now an explicit function representation, mapping the state, $\vectorstyle{s}_t$, to the corresponding optimal action, $\vectorstyle{a}_t$, it is possible to expand the input space with additional features that may encode useful information to build the highly nonlinear mapping. Therefore, as we want to implement the dependency on the changing goal, we include the goal information in the state, resulting in an augmented state and a contextual policy.
The previous robot that performed the action is included as well, this to nudge the policy in the direction of switching between robots.
\begin{equation}
	\label{eq:state}
	\hat{\vectorstyle{s}}_t = (x_t,y_t,\theta_t,x_{g},y_{g},\theta_{g},b_{t-1}) \in \mathcal{S}^2\times \{\text{L},\text{R}\}
\end{equation}
Two different policy architectures are considered. One where the policy defines the final step and one where this final step is determined externally, using the kinematics of both manipulators.

\paragraph{Architecture A}
One of the main complexities in the problem arises from performing the final step. The final action is defined by the Boolean $f_t$ and affects the problem in a highly nonlinear fashion. When eliminating this Boolean from the action, after each step it needs to be checked if the object in the mutual manipulation space corresponds with the desired goal configuration. As the complete goal information is provided in the augmented state, this check can be performed for both manipulators. This check is performed by placing the block at the goal position in simulation and check if the desired orientation is achieved. If this orientation is achieved for one of the two manipulators the object performs the final step.

Because the final step is no longer dictated by the policy the action space reduces to
$$\hat{\vectorstyle{a}}_t = (b_t,\alpha_t,\beta_t)\in \mathcal{A}'=\{\text{L},\text{R}\}\times[0,1]^2$$
where $\alpha_t$ and $\beta_t$ represent $x_{t+1}$ and $y_{t+1}$ respectively.

The policy now has to find a sequence of actions resulting in a state in the mutual manipulation space that correspond to the goal configuration. The policy, however, has no idea what robot performs the final step as this is dictated externally.

\paragraph{Architecture B}
If the policy dictates the final step, it has to determine which manipulator needs to perform it and when it needs to performed. It has to dictate which state in the mutual manipulation space corresponds to which goal state for each possible manipulator, while also deciding the sequence of actions to get to that state. The final step is now a part of the action sequence. The state is unaltered from equation \ref{eq:state}, while action \ref{eq:action} is adapted to include the continuous parametrization.

$$\hat{\vectorstyle{a}}_t = (f_t,b_t,\alpha_t,\beta_t)\in \mathcal{A}'=\{0,1\}\times\{\text{L},\text{R}\}\times[0,1]^2$$



\subsubsection{\textbf{Reinforcement learning}}
We learn this policy using reinforcement learning. The solution space is parametrized using continuous variables, therefore it is opted to go with an actor-critic method, that concurrently learns a Q-function and a policy.

\paragraph{Learning algorithm}
We opted to go for Soft Actor Critic, as this is a popular choice and provided a stable implementation. Although other methods could have worked as well, such as TD3 \cite{Fujimoto2018} or PPO \cite{Schulman2015}. SAC optimizes a stochastic policy in an off-policy way, with as main feature the use of entropy regularization. It is trained to maximize a trade-off between expected return and entropy. Entropy is a measure of the randomness in the system. If $x$ is a random value, with a probability mass $P$. The entropy $H$ of $x$ can be distributed from the its distribution $P$ according to
$$H(P) = \underset{x \sim P}{\mathbb{E}}(-\log P(x))$$
The agent will get a bonus reward each time step proportional to the entropy of the policy at that timestep. Increasing entropy results in more exploration, which can consequently improve learning later on. Additionally, converging to a bad local optimum can be prevented. The problem description changes to
$$\pi^* = \arg \max_{\pi} \underset{\tau \sim \pi}{\mathbb{E}}\sum_{t=0}^{\infty} \gamma^t \big( R(s_t, a_t, s_{t+1}) + \alpha H\left(\pi(\cdot|s_t)\right) \big)$$
Further details can be found in the corresponding literature \cite{Haarnoja2018}.

\paragraph{Reward shaping} 
In order to minimize manual tuning we opted for sparse rewards for both architectures, with no extra information being encoded in the reward. For the first architecture this is rather easy, since we have only two scenarios. If the state of the object in the mutual workspace correspond to a desired goal state for one of the two manipulators, the episode is terminated and the object is placed at its goal, without receiving a negative reward. In the other case, where the state is not as desired, the episode continues and a reward of minus one is returned. This way the episode length is minimized.
$$r_A = \begin{cases}
0, & |\theta_t^{b_{t+1}} - \theta_{g}| \leq \epsilon  \\
-1, & |\theta_t^{b_{t+1}} - \theta_{g}| > \epsilon
\end{cases}$$
with $\theta_t^{b_{t+1}}$ indicating the orientation at the goal position, placed there by robot $b_{t+1}$.

For architecture B however new ways for an episode to fail are introduced. In architecture B, since the policy dictates when to place the object at the desired position, the object can be placed at the final position at the wrong time. Additionally, the policy can query the wrong robot, resulting in either a wrong orientation or an infeasible goal position. Finally, the policy could never make the decision to take the final step.

The following sparse reward function is used
$$r_B =
\begin{cases}
20, & (x_t,y_t) \equiv (x_g,y_g) \text{ and } \ |\theta-\theta_{g}| \leq \epsilon \\
-1, & (x_t,y_t) \in \mathcal{M}_m \\
-30, & \quad (x_t,y_t)\equiv (x_g,y_g) \text{ and } \ |\theta-\theta_{g}| > \epsilon \\
	& \text{or} \quad (x_t,y_t) \lor (x_{t-1},y_{t-1}) \notin \mathcal{M}_{b_t}
\end{cases}$$

with $\mathcal{M}_{b_t}$ indicating the workspace of the robot performing the action at $t$. The final case indicates when the robot needs to pick-up the object outside its manipulation space, which is infeasible. To avoid ending up in a local minimum, where the object is placed directly at the goal, resulting in a reward of -1, a positive reward is given if the goal is achieved. Additionally, a strong negative reward is returned for infeasible actions.

\paragraph{Hindsight experience replay}
Since we are dealing with sparse rewards, when the block is misplaced the reward provides no additional information except that the performed sequence of actions does not lead to a successful goal configuration. It is however possible to act as if the found configuration was the desired goal, which gives us extra information. This is exactly what Hindsight Experience Replay does , for the specific details we refer to \cite{Andrychowicz2017}.

\paragraph{Further remarks}
SAC is used to train both architectures. In the first the action has dimension 3, while the action has four dimensions in architecture B. The latter is harder to learn, therefore a simplified form of curriculum learning \cite{Bengio2009} is implemented. The policy is first trained on a simpler environment where the allowed mistake on the orientation is larger. This gives the policy an incentive to move to the goal position faster, which helps training. By gradually increasing the difficulty by making the allowed error smaller, the policy learns eventually in the correct environment. In practice two training steps are enough in order to get to the desired accuracy.

In order to train the policy a significant amount of episodes need to be performed. Training a large number of episodes on a real setup would be impractical because it would take too long, but the deterministic nature of the problem allow us to use simulation instead. This is because the policy can be trained without taking motion planning into account, since the state update rule does not depend on the motion plan. We only need to calculate the robot poses at the drop-off and pick-up locations. The trained policy in simulation will then be executed on the real setup. However, the accuracy of this translation from simulated to real environment depends on the accuracy of the models and the repeatability of the two robotic manipulators. The closed-loop behavior of the supervisory controller can help mitigate some of these issues.

\subsection{Related work} \label{sec:previous}
A previous method presented by De Witte et al. \cite{DeWitte2022} uses the discretized version of the workspace, instead of using a continuous parametrization. A policy is found using DQN \cite{Mnih2015}, where the optimal action-value function is approximated using neural networks $Q(\vectorstyle{s},\vectorstyle{a};\theta) \approx Q^*(\vectorstyle{s},\vectorstyle{a})$. The best action is then found by taking the action with maximum value $\pi(\vectorstyle{s}) = \max_{a} Q(s,a;\vectorstyle{\phi})$. Discretizing this shared workspace results in a high-dimensional solution space that is hard to solve, resulting in longer training times. 

Additionally, for architecture B an extensive reward shaping and extensive and complex pretraining, with multiple networks, are required for architecture B. Architecture A on the other hand required two policies, due to how the information of the goal configuration was implemented. Moreover, the policy is only trained to output a certain set of discrete possible positions. If this is applied on a real set-up, where the execution of actions is no longer deterministic, the actual achieved position may differ, resulting in reduced accuracy of the goal task as this position is not yet seen before by the policy. All of these problems are mostly mitigated in our approach.

\section{Experimental Validation}
\subsection{Set-up}
The set-up uses two Lynxmotion AL5A robotic arms with five degrees of freedom to perform pick-and-place maneuvers on a square object by grasping a toothpick attached at its center. The robots are controlled in an open-loop manner, moving from the pick-up position to the drop-off position. An Intel RealSense Depth Camera D435 is used to perceive information from the system, such as the configuration of the block and the goal.

The current path planning method uses a parametric trajectory optimization. The trajectory $\zeta$ is parametrized in joint coordinates using interpolating B-splines. The number of knots $m$ and the degree $d$ determine the number of splines, $N = m+d-1$. In our experiments $(m, d)$ are chosen as $(3,5)$. As the problem is purely kinematic and dynamics are ignored, time parametrization is of no importance and we parametrize time with an arbitrary variable $\tau \in [0,1]$.
$$\zeta(\tau;\theta) = \sum_{i=1}^N B_{i,d}(\tau)\theta_i$$

Collision avoidance is considered for both the ground and the robots themselves, with the robots being limited in the mutual workspace to prevent collisions with each other. The position in Cartesian coordinates of all the links is evaluated using $10^2$ intervals. The optimization minimizes the distance in Cartesian coordinates, with the integral being calculated with trapezoidal integration.

The path planning can be implemented using any existing trajectory optimization or motion planning method, it is not limited to the one used in our approach, due to the decoupling of the task and motion planning.

\subsection{Implementation}
All code is written in Python. The reinforcement learning algorithms were implemented using the stable baselines 3 package \cite{stable-baselines3}.

\subsubsection{Continuous approximation}
\begin{figure}
	\centering
	\includegraphics[width=\columnwidth]{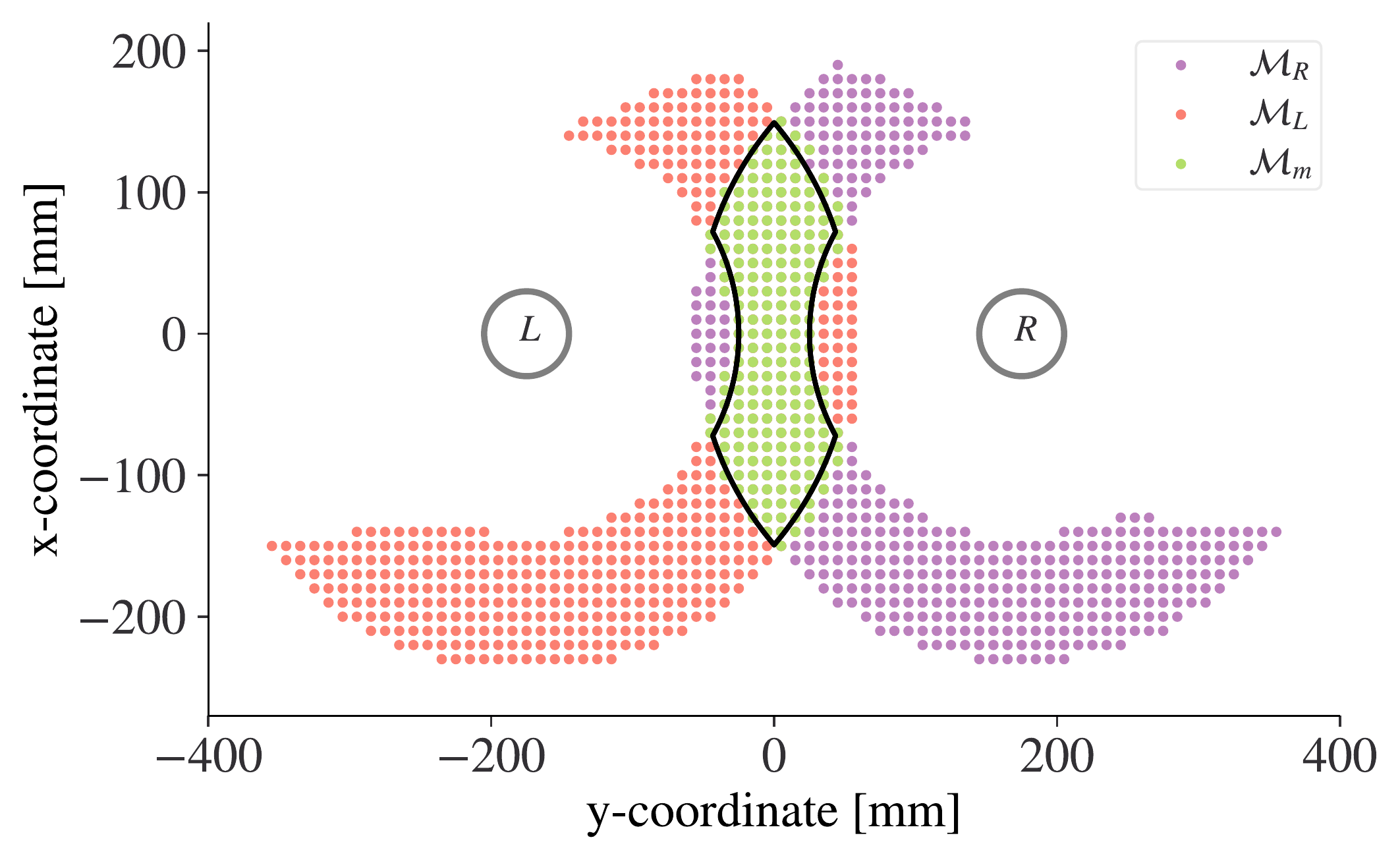}
	\caption{Continuous approximation shown on top of discrete approximations.}
	\label{fig:cont_approx}
\end{figure}

A two-dimensional grid of points at a fixed height, spaced apart 5 mm in each direction, is sampled. For each point the feasibility is checked for both robots, resulting in a discrete approximation of the manipulation spaces. Taking the intersection results in a discrete approximation of the mutual manipulation space, which can be used to find the continuous approximation. The resulting $\underline{r}$ and $\overline{r}$, are 150 and 230 respectively. The goal and initial configuration positions are sampled from the union of the two discrete manipulation spaces. These discrete approximations, together with the continuous approximation in black, are given in Fig. \ref{fig:cont_approx}. The base of both robots are indicated as well.

\subsubsection{Hyperparameter tuning}
An extensive hyperparameter search is performed for all different architectures using Optuna \cite{optuna_2019}. The resulting hyperparameters for all continuous cases can be found in table \ref{tab:sac_hyperparams}. ReLU is used as activation function and Adam \cite{Kingma2017} as optimizer. Architecture B is trained using HER, with five goals being generated from all states of an episode. For the first architecture HER seems to provide no added benefit, which is to be expected. The maximum allowed episode length is twenty, while taking a random policy results in an average length of around 10. This means that using HER results in completed episodes of length twenty being added to the replay buffer, which is rather long in comparison to a random policy. When limiting this maximum episode length to \eg five, HER could help. The second architecture clearly benefits from using HER, on which we will elaborate in the results section. Additionally, since the final step is not dictated by the policy, the goals can only be generated from the final state of the episode, while for architecture B it is possible to generate the goals from all states in the episode.

\begin{table}[htbp]
	\caption{Hyperparameters SAC for both architectures.}
	\label{tab:sac_hyperparams}
	\centering
	\begin{tabular}{ccc}
	  	\toprule
	  	\textbf{Hyperparameter} & \textbf{Architecture A} & \textbf{Architecture B} \\
	  	\midrule
		learning rate & 0.0003 & 0.0005 \\
		discount ($\gamma$) & 0.99 & 0.99 \\
		replay buffer size & $10^5$ & $10^5$ \\
		batch size & 32 & 128 \\
		learning starts & 0 & 1000 \\
		train frequency & 32 & 16 \\
		target smoothing coefficient ($\tau$) & 0.01 & 0.01 \\
		number of hidden layers & 2 & 3 \\
		number of hidden units per layer & $400|300$ & 512 \\
		\bottomrule
	\end{tabular}
\end{table}

\subsection{Results}
The results are compared to the method presented in \ref{sec:previous}. The main improvement can be found in the ease of training and the higher success rate for architecture B.

\subsubsection{Training}
Training is performed with the optimal hyperparameters mentioned previously. 

A first comparison is made between the simplified form of curriculum learning and learning directly in the final environment. It can be seen, that while training seems to move towards an optimal point, it does so slower than the simplified form. This gives a validation for the use of the curriculum learning, as training speed is increased. On the other hand, it also shows that curriculum learning is not required and a policy can be found without it. Here, the first clear benefit of the method over the discretized method surfaces, training is much simpler. On top of being simpler, it is significantly faster for architecture B (2,150,000 vs 650,000 steps).

\begin{figure}
	\centering
	\includegraphics[width=\columnwidth]{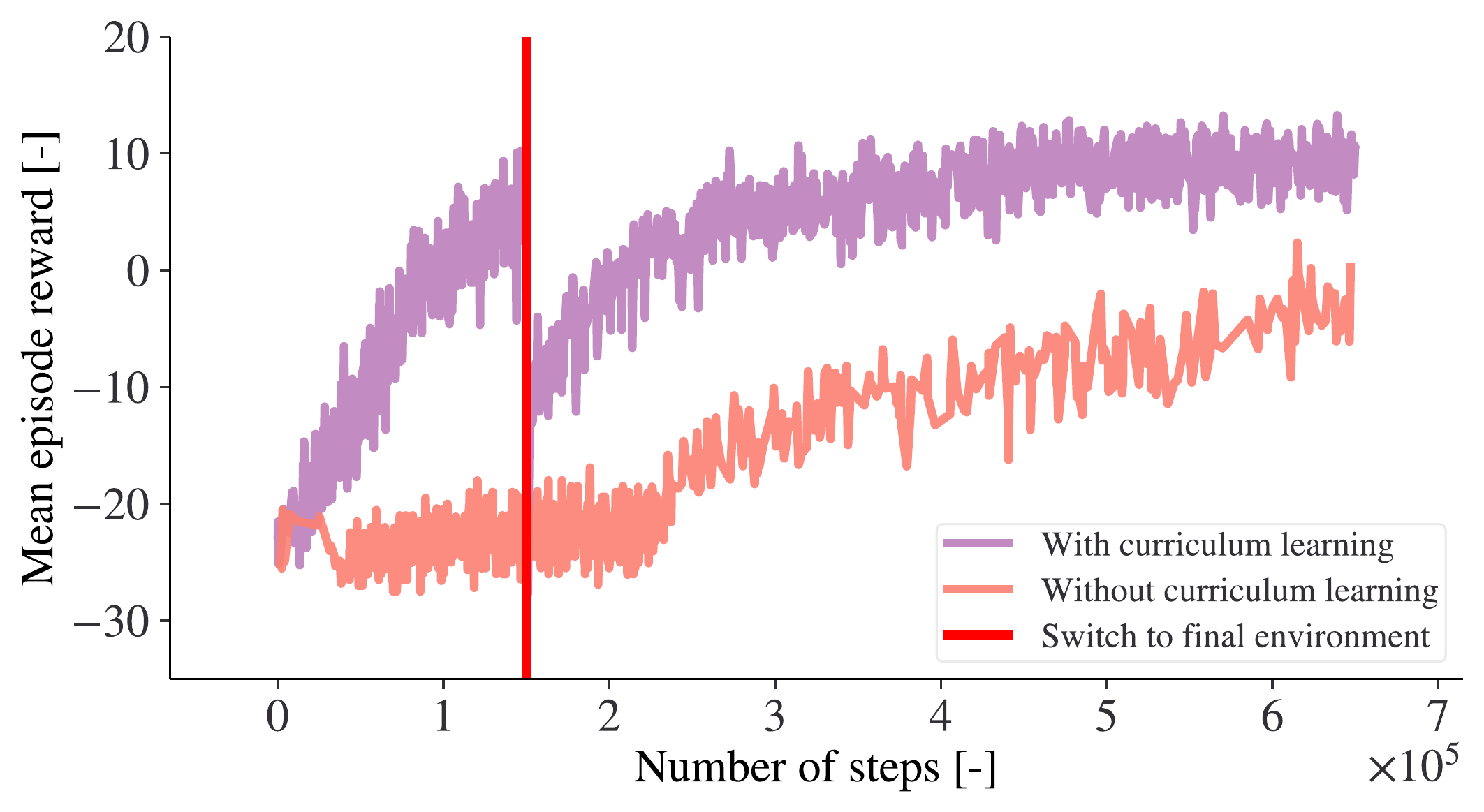}
	\caption{A comparison made between curriculum learning and no curriculum learning, the switch of environments for curriculum learning is indicated at step 150,000.}
    \label{fig:curric_compar}
\end{figure}

Secondly we can take a look at the mean episode length during training for both methods. The main difference is the use of HER, training architecture B is done using HER, which can be seen by the sudden decrease in mean episode length. As shorter episodes that reach their goal are created. Additionally, since the policy dictates when an episode is done, even a bad policy will result in shorter episodes.

\begin{figure}
	\centering
	\includegraphics[width=\columnwidth]{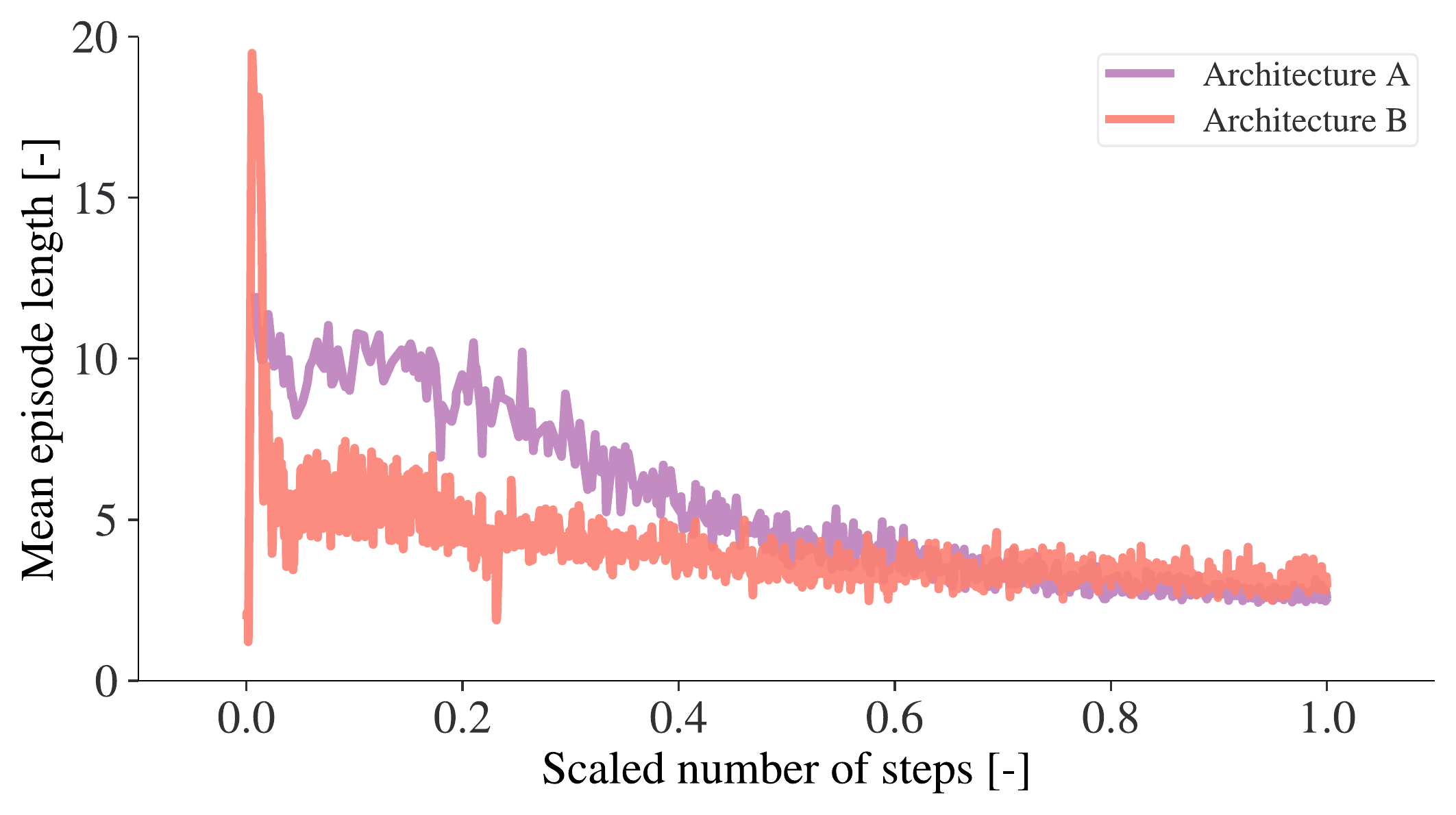}
	\caption{A comparison made of the mean episode length during training for both architectures. Architecture A is trained for 250,000, while architecture B is trained for 650,000. The number of steps are scaled for a better comparison.}
    \label{fig:arch_compar}
\end{figure}

It can be noted that uniform sampling of $\alpha$ and $\beta$ will result in a non-uniform distribution of points in Cartesian coordinates, because of the changing width. While this seems suboptimal at first sight, the non-uniform sampling may benefit the exploration phase. The positions with a higher density lie on the edges of the mutual manipulation space. Switches between edges results in a higher change of relative orientation, thus benefitting the exploration.

\subsubsection{Validation}
We perform a number of experiments with a randomly sampled goal and start state for all four architectures, both in simulation as on the set-up. The average episode length and the success rate are given. The success rate is given as the percentage of episodes that resulted in a desired goal; 50 episodes are performed.
 
\begin{figure}
	\centering
	\includegraphics[width=\columnwidth]{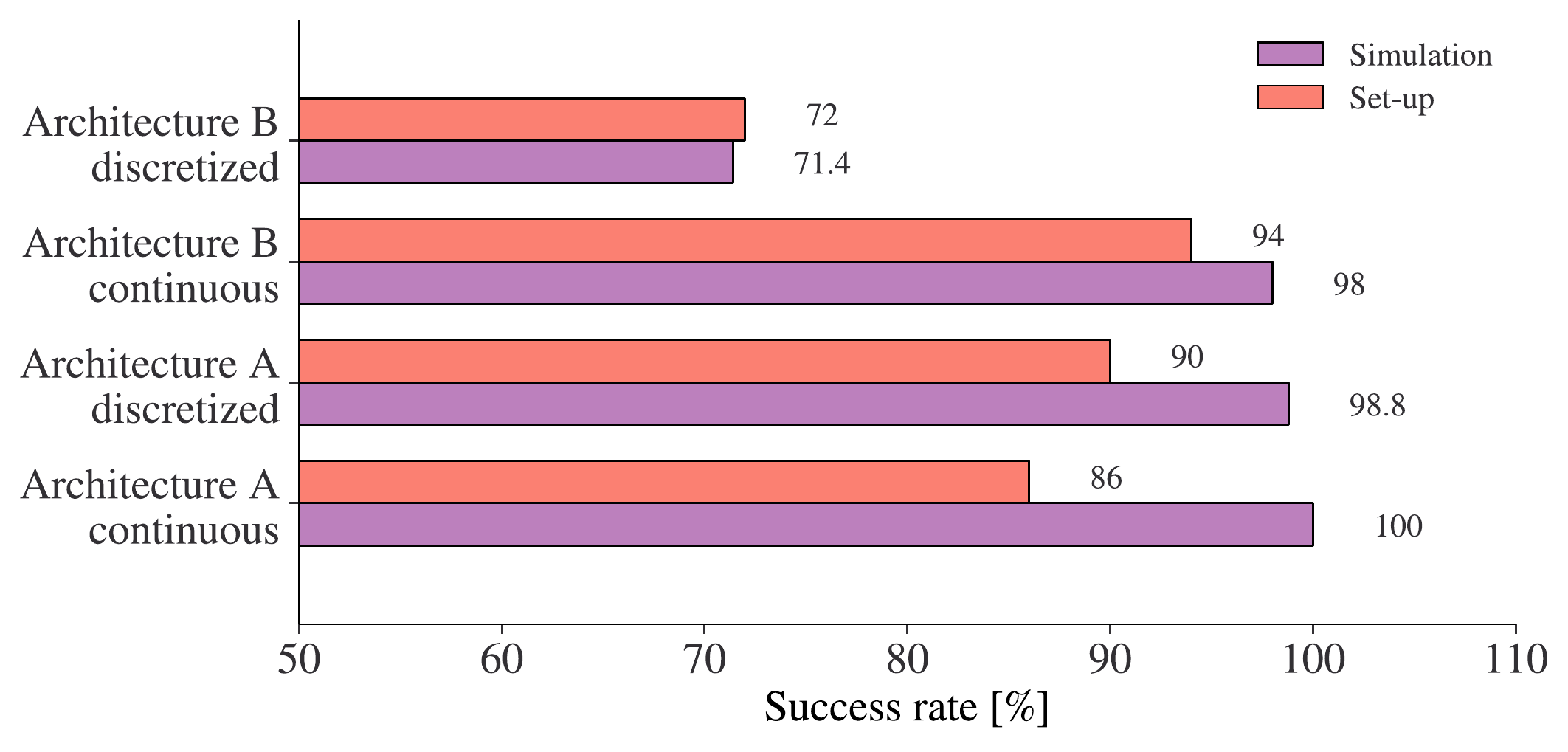}
	\caption{A comparison made between the discretized version and the continuous version of the shared workspace, both for architecture A and B. The success ratios are shown both on the set-up as in simulation.}
    \label{fig:bar_sr}
\end{figure}

\begin{figure}
	\centering
	\includegraphics[width=\columnwidth]{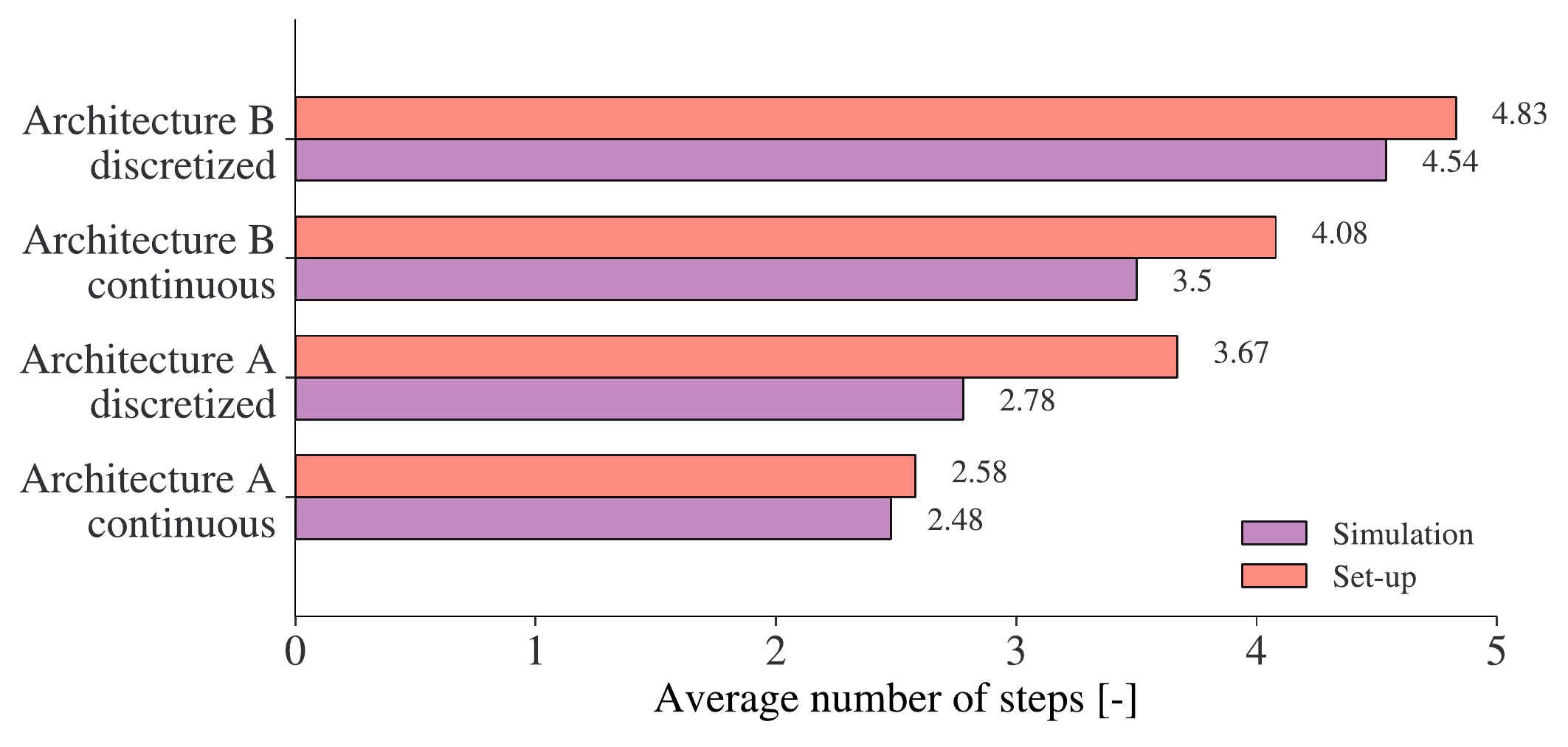}
	\caption{A comparison made between the discretized version and the continuous version of the shared workspace, both for architecture A and B. The average number of steps are shown both on the set-up as in simulation.}
    \label{fig:bar_steps}
\end{figure}

The results in simulation and on the setup are given in Fig. \ref{fig:bar_sr} and \ref{fig:bar_steps}. A comparison is made with the discretized version of the shared workspace for both architectures. The first figure shows the success rate, while the second shows the average number of steps in an episode. A clear improvement can be seen when using the continuous representation over the discretized, especially for architecture B, both on the setup as in simulation and both in episode length as in success rate. While the same can be said for architecture A and its episode length, the success rate is more or less the same. It shows that for the simpler problem statement this method doesn't provide a massive boost in performance, while for the harder problem statement the parametrization of the mutual workspace using continuous variables is almost required.

For training in simulation no physics engine was required, since only the initial $x_t$ and final state $x_{t+1}$ of the motion plan where needed. However for the evaluation a physics engine is used to be as complete as possible. The robots' dynamics and interaction with the environment are simulated using PyBullet. By modeling the robot in a physics engine, we are able to better validate the trained policies.

For architecture B, the possible end state are: (a) Correct position and orientation, (b) correct position and wrong orientation, (c) wrong position. The wrong position can be achieved in two ways, if the wrong robot is asked to perform the final step it could be possible the position lies outside the manipulation space of that robot, or if the max amount of steps is achieved. Of the 50 performed episodes two resulted in the final state (b) and one in (c) by querying the wrong robot. For architecture A, all episodes where terminated before maximum number of steps, so the low success rate is due to wrong model of the robot. The latter is less of an issue for architecture B, since the model is not used to dictate the final step, resulting in a better translation to the set-up.

\section{Conclusion}
This work presents a method for enabling multi-robot manipulation in the presence of under-actuation by utilizing the shared workspace. The proposed approach involves decomposing the resulting Task and Motion Planning problem and using a supervisory control system to output the next best action in real-time. The next best action is selected from a continuous parametrization of the shared workspace by a parametrized policy. This continuous approximation results in a simple to train policy, using only sparse rewards, with a shorter training time and a higher success rate than previous methods. A translation to a real set-up shows it's application in the real-world. 

The parametrization of actions into continuous variables can be extended to other problems as well, enabling us to find an action sequence by learning a policy. An other example could be to parametrize the workspace of a single robot, and combine this with a number of individual actions, which could be parametrized in a similar way. Cooperation could also be performed by parametrization of a three-dimensional workspace, making it for example possible to perform hand-over actions.


In future work, it may be possible to extend the system to include more robots or a greater range of possible actions, and to consider different types of goal configurations. Another step could be to implement the method with a combined Task and Motion Planning solver, resulting in a faster combined optimization. Additionally, it could be possible to give the policy more information of its environment to make it even more contextual. A variable representing obstacles could be included in the augmented state, such that the policy can redirect its next position.

\bibliography{refs}{}
\bibliographystyle{IEEEtran}

\end{document}